\documentclass[10pt,conference]{IEEEtran}
\IEEEoverridecommandlockouts
\usepackage{cite}
\usepackage{amsmath,amssymb,amsfonts}
\usepackage{algorithmic}
\usepackage{graphicx}
\usepackage{textcomp}
\usepackage{color}
\usepackage[dvipsnames]{xcolor}
\usepackage[T1]{fontenc}
\def\BibTeX{{\rm B\kern-.05em{\sc i\kern-.025em b}\kern-.08em
    T\kern-.1667em\lower.7ex\hbox{E}\kern-.125emX}}

\usepackage{hyperref}
\usepackage{booktabs}
\usepackage{multirow}
\usepackage{multicol}
\usepackage{array}

\usepackage[utf8x]{inputenc}
\usepackage[linesnumbered,ruled,vlined]{algorithm2e}
\SetAlFnt{\small}

\usepackage{etoolbox} 
\usepackage{orcidlink}
\usepackage{marvosym} 
\usepackage{soul} 
\sethlcolor{SpringGreen}
\usepackage{tabularx}

\usepackage{titlesec}
\titlespacing{\section}{0pt}{6pt}{4pt}
\titlespacing{\subsection}{0pt}{4pt}{2pt}

\begin{document}

\title{
Surrogate-Assisted Genetic Programming with Rank-Based Phenotypic Characterisation for Dynamic Multi-Mode Project Scheduling
}

\author{\IEEEauthorblockN{Yuan Tian
                        \thanks{
                        This work has been submitted to the IEEE for possible publication. Copyright may be transferred without notice, after which this version may no longer be accessible.
                        }
                        \textsuperscript{\Letter},
                        Yi Mei,
                        Mengjie Zhang
                        }
\IEEEauthorblockA{\textit{Centre for Data Science and Artificial Intelligence \& School of Engineering and Computer Science} \\
\textit{Victoria University of Wellington, PO Box 600, Wellington 6140, New Zealand}\\
\{yuan.tian, yi.mei, mengjie.zhang\}@ecs.vuw.ac.nz}
}


\IEEEoverridecommandlockouts


\maketitle


\IEEEpubidadjcol

\begin{abstract}

The dynamic multi-mode resource-constrained project scheduling problem (DMRCPSP) is of practical importance, as it requires making real-time decisions under changing project states and resource availability. Genetic Programming (GP) has been shown to effectively evolve heuristic rules for such decision-making tasks; however, the evolutionary process typically relies on a large number of simulation-based fitness evaluations, resulting in high computational cost. Surrogate models offer a promising solution to reduce evaluation cost, but their application to GP requires problem-specific phenotypic characterisation (PC) schemes of heuristic rules. There is currently a lack of suitable PC schemes for GP applied to DMRCPSP. This paper proposes a rank-based PC scheme derived from heuristic-driven ordering of eligible activity–mode pairs and activity groups in decision situations. The resulting PC vectors enable a surrogate model to estimate the fitness of unevaluated GP individuals. Based on this scheme, a surrogate-assisted GP algorithm is developed.
Experimental results demonstrate that the proposed surrogate-assisted GP can identify high-quality heuristic rules consistently earlier than the state-of-the-art GP approach for DMRCPSP, while introducing only marginal computational overhead. Further analyses demonstrate that the surrogate model provides useful guidance for offspring selection, leading to improved evolutionary efficiency.
 
\end{abstract}


\section{Introduction}

Project scheduling is a fundamental component of project management, as high-quality scheduling decisions are essential for completing projects within budget and time constraints. The multi-mode resource-constrained project scheduling problem (MRCPSP) \cite{peteghem_experimental_2014, hartmann_updated_2022} is a well-formalised problem setting, in which the start times and execution modes of activities must be determined under precedence and resource constraints to minimise the overall duration of the project. In the dynamic MRCPSP (DMRCPSP) considered in this study, the actual durations of activities are uncertain and only become known during execution. As a result, scheduling decisions—specifically, which activities should be executed and in which modes—must be made online as the project state updates. 

Heuristic rules are well-suited to such dynamic decision-making scenarios due to their low computational cost. Genetic Programming (GP) \cite{zhang_survey_2023}, as a hyper-heuristic approach, has been widely used to automatically evolve scheduling heuristics based on problem-specific attributes and decision contexts. However, evolving effective heuristics requires a large number of expensive simulation-based fitness evaluations, which severely limit scalability and hinder practical applicability in dynamic decision-making environments.

Surrogate-assisted GP \cite{hildebrandt_using_2015, zhang_surrogate-assisted_2021} has shown promise in reducing evaluation costs for scheduling heuristics. However, its successful application critically depends on the availability of problem-specific phenotypic characterisation (PC) schemes that can meaningfully capture the behavioural differences among heuristics. For DMRCPSP, there does not exist such a scheme, and the existing schemes in related problems are not applicable in this case, which prevents the surrogate-assisted GP frameworks from being directly applied to this problem.

Motivated by the need to enable computationally efficient GP-based decision-making for DMRCPSP, this paper aims to bridge the above gap by developing a suitable phenotypic characterisation and integrating it into a surrogate-assisted GP approach. Specifically, the research objectives of this work are as follows:

\begin{itemize}
    \item To design a problem-specific phenotypic characterisation scheme for GP heuristics in DMRCPSP based on the ranking behaviour of heuristics over eligible activities.
    \item To develop a surrogate-assisted GP algorithm that incorporates the proposed PC scheme.
    \item To verify the effectiveness of the PC scheme and compare the performance of the surrogate-assisted GP with the state-of-the-art GP for DMRCPSP.
\end{itemize}

\section{Background}
\subsection{Problem Description}
The dynamic multi-mode resource-constrained project scheduling problem considers the scheduling of a set of project activities subject to precedence and resource constraints. Each activity can only start after all of its predecessor activities have been completed. For each activity, multiple execution modes are available, each with a different expected duration and resource requirement.
In a dynamic environment, the exact duration of an activity is not known in advance and is only revealed during execution. The actual duration varies within a predefined range determined by optimistic and pessimistic estimates around the expected duration. At any point in time, only a limited amount of renewable resources is available, and the total resource consumption of activities executed in parallel must not exceed the corresponding resource capacities.
The objective of the DMRCPSP is to construct a feasible project schedule that respects both precedence and resource constraints while minimising the overall project makespan.
\subsection{Genetic Programming for DMRCPSP}
Genetic Programming (GP) has been widely applied as a hyper-heuristic method for solving complex scheduling problems, including standard RCPSP \cite{dumic_ensembles_2021,luo_efficient_2022,umic_using_2022}, job shop scheduling\cite{zhang_survey_2023}, and related domains\cite{wang_explainable_2023,guo_genetic_2024}. In GP-based scheduling, heuristic rules are evolved to guide decision-making during schedule construction or simulation.

For the dynamic multi-mode RCPSP, recent studies \cite{tian_learning_2024,tian_genetic_2025,tian_scalable_2025} have focused on improving decision-making mechanisms to efficiently utilise resources. A representative approach \cite{tian_scalable_2025} formulates schedules to utilise resources efficiently using heuristic rules. First, an activity ordering rule ranks eligible activity–mode pairs according to their priority values, and a promising subset of candidates is selected. Next, feasible activity groups are enumerated from the selected candidates, and an activity group selection rule determines which group should be executed. GP employs a multi-tree representation to evolve these two rules simultaneously.

This decision-making framework is effective for DMRCPSP, as it allows GP to jointly optimise activity ordering and group selection strategies. However, the computational cost of evaluating GP individuals remains a major bottleneck, since each individual must be embedded into a simulation to assess its scheduling performance.

\subsection{Surrogate Models for Scheduling Heuristics}
Surrogate models \cite{jin_data-driven_2021} are a class of techniques designed to accelerate evolutionary algorithms on such problems by approximating expensive fitness evaluations using computationally cheaper alternatives. The basic idea is to replace a portion of full evaluations with inexpensive estimations based on previously evaluated solutions. 

In the design of surrogate models for GP-based scheduling heuristics, two main approaches have been explored in the literature: simplified simulation models \cite{nguyen_surrogate-assisted_2017} and phenotypic characterisation-based methods  \cite{hildebrandt_using_2015}.
The first approach relies on replacing the original, computationally expensive evaluation with a simplified simulation model, for example, by reducing the number of jobs, machines, or resources considered during evaluation. While such simplified models can reduce the computational cost to some extent, the time savings are often limited. Moreover, their effectiveness depends on carefully selecting which problem parameters can be reduced without significantly compromising evaluation fidelity, which itself requires additional empirical investigation.

The second approach adopts a more abstract and computationally efficient strategy. Instead of executing a full (or reduced) simulation, this approach extracts a set of representative decision situations from the original simulation environment. GP heuristics are then applied to these situations to make decisions, and their decision-making behaviours are recorded and transformed into numerical feature vectors. As a representative example, reference  \cite{hildebrandt_using_2015} proposed a phenotypic characterisation scheme for the job shop scheduling problem based on a reference rule. In each decision situation, all candidate operations are ranked and indexed according to a predefined reference rule, and the operation selected by a GP heuristic is represented by its corresponding rank index. Consequently, the phenotypic characterisation (PC) vector of a GP individual consists of the indices of the selected operations across all decision situations.
These PC vectors can then be used to measure behavioural similarity between GP individuals, and, when combined with known fitness values, to estimate the fitness of unevaluated individuals using surrogate models such as nearest-neighbour regression. A key challenge in applying PC-based surrogate methods lies in the design of an effective scheme that can faithfully map the behaviour of GP heuristics to a numerical vector representation.

The PC scheme introduced in \cite{hildebrandt_using_2015} has subsequently been applied to other scheduling domains, including flexible job shop scheduling \cite{zhang_surrogate-assisted_2021} and multi-project RCPSP \cite{chen_surrogate-assisted_2023}, but this scheme does not apply to GP for DMRCPSP. The main limitation lies in the fact that the scheme in \cite{hildebrandt_using_2015} is designed for single-choice decision scenarios, whereas the heuristic rules in \cite{tian_scalable_2025} are applied to both ranking-based decision situations and activity group selection, which corresponds to subset selection. While subset selection can be adapted by enumerating and indexing activity groups, enumerating activity orderings is impractical due to the factorial growth of permutations. Moreover, assigning indices to permutations destroys the semantic relationships between different orderings. In addition, the PC scheme in \cite{hildebrandt_using_2015} records only the rank of the highest-priority candidate determined by a reference rule, ignoring the relative ordering of other candidates in the same decision situation. As a result, two heuristic rules that select the same highest-priority candidate but differ substantially in the ordering of remaining candidates may be considered identical, leading to inaccurate distance measurements between GP individuals.

These limitations motivate the development of a new PC scheme that can effectively capture ranking-based and subset-selection decision behaviours in DMRCPSP, enabling the application of surrogate-assisted GP to this problem.

\section{Proposed Method}
\subsection{Overall Framework}

\begin{figure}
    \centering
    \includegraphics[width=\linewidth]{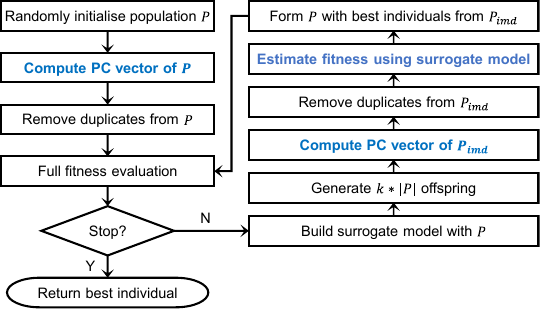}
    \caption{Flowchart of the surrogate-assisted GP algorithm.}
    \label{fig:SGP_flowchart}
\end{figure}

The proposed algorithm adopts the framework from \cite{hildebrandt_using_2015}, representing a conventional surrogate-assisted GP workflow. An overview of the algorithm is illustrated in Fig. \ref{fig:SGP_flowchart}, where the differences introduced in this work are highlighted in blue.

The algorithm begins with an initial population $P$. For each individual in the population, a phenotypic characterisation (PC) vector is computed; the scheme of PC is detailed in Section \ref{section:PC_scheme}. Duplicate individuals in terms of PC vector are then removed, and the remaining unique individuals are evaluated using the full fitness function. During the evolutionary loop, the PC vectors of the current population are used to construct a surrogate model. Subsequently, $k\times|P|$ intermediate offspring $P_{imd}$ are generated, where $k$ is referred to as the offspring multiplier. The PC vectors of the intermediate offspring are computed, the offspring with the unique vector are kept, and the surrogate model is employed to estimate their fitness values (details of the surrogate estimation are provided in Section \ref{section:fitness_prediction}). Finally, the top $|P|$ offspring, as ranked by the surrogate, are selected for full fitness evaluation. The evolutionary loop repeats until the stopping criterion is met, and the best individual found is returned.

Since this work extends the approach in \cite{tian_scalable_2025} by introducing a surrogate model, GP individual representation, full fitness evaluation procedure, as well as crossover, mutation, and selection operators, remain the same as in that work; thus, they are omitted in this paper. More details can be found in \cite{tian_scalable_2025}.

\subsection{New Phenotypic Characterisation Scheme}
\label{section:PC_scheme}

A well-designed PC scheme should reflect the behaviour of GP individuals during full fitness evaluation. In this work, each individual consists of two trees: an activity ordering rule and an activity group selection rule. The activity ordering rule is applied to compute ordering priority values for eligible activity–mode pairs at each decision point. Based on these priority values, a knee point selection strategy is employed to identify a promising subset of activity–mode pairs. From this subset, feasible activity groups are enumerated, and the activity group selection rule is then used to select the activity group with the highest priority for execution.

\begin{figure}
    \centering
    \includegraphics[width=\linewidth]{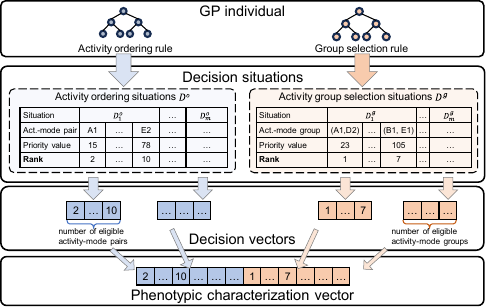}
    \caption{Phenotypic characterisation scheme of a GP individual.}
    \label{fig:rank_diff_scheme}
\end{figure}

Motivated by the distinct roles of the two heuristic rules, we propose a decision-situation-based PC scheme, as illustrated in Fig. \ref{fig:rank_diff_scheme}. Two types of decision situations are considered: activity ordering situations and activity group selection situations. In each decision situation, the eligible activities and the corresponding project state information are obtained from data sampled during the schedule simulation.

For a given GP individual, the corresponding heuristic rule is applied to each decision situation to compute priority values and rankings for all eligible activity–mode pairs or activity–mode groups. A smaller priority value indicates a higher rank. The ranks of all candidates in each decision situation are extracted to form a decision vector. Finally, the decision vectors obtained from all decision situations are concatenated to produce the complete PC vector of a GP individual. This scheme captures the ranking behaviour of heuristic rules across representative decision contexts, providing a phenotypic description of GP individuals.

\subsection{Surrogate-Based Fitness Estimation}
\label{section:fitness_prediction}

The surrogate model is used to predict the relative quality of candidate offspring for selection purposes, rather than to replace the true fitness evaluation. In this work, a simple nearest-neighbour surrogate model is employed for fitness estimation. The surrogate takes as input the PC vectors of the current population along with their corresponding fitness values. 
To estimate the fitness of a new individual, the Manhattan distance is computed between its PC vector and the PC vectors of all individuals stored in the surrogate database. The fitness value of the nearest neighbour in the PC space is then assigned as the estimated fitness of the new individual. It is worth noting that the training samples in the surrogate model are not accumulated across generations. Since the full fitness evaluation involves stochastic simulation with different random seeds in each generation, fitness values of individuals from different generations are not directly comparable. Therefore, the surrogate database contains only individuals from the current population.

\begin{figure}
    \centering
    \includegraphics[width=\linewidth]{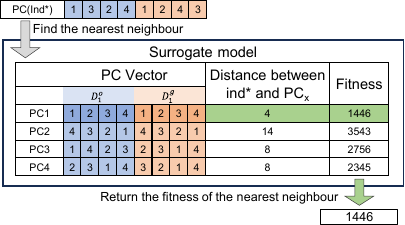}
    \caption{Example of fitness estimation by the surrogate model.}
    \label{fig:fitness_prediction}
\end{figure}

An illustrative example of the fitness estimation process is shown in Fig. \ref{fig:fitness_prediction}. In this example, the PC vector consists of one activity ordering situation and one activity group selection situation. The surrogate database contains the PC vectors and fitness values of four individuals. When predicting the fitness of a new individual, denoted as $Ind^*$, the Manhattan distances between $Ind^*$ and the four PC vectors are computed. $Ind^*$ is closest to $PC1$ with a distance of 4; the fitness value associated with $PC1$ is returned as the estimated fitness.

\section{Experimental Study}

\subsection{Experiment Deign}
To evaluate the effectiveness of the proposed approach, the algorithm in \cite{tian_scalable_2025}, namely KGGP, is adopted as the baseline for comparison. All algorithms use the same simulation environment and GP parameter settings as in \cite{tian_scalable_2025} to ensure a fair comparison. In the simulation, each project consists of 200 activities, where each activity can be executed in three modes, and a total of 12 renewable resource types are considered. The precedence complexity among activities is characterised using the order strength \cite{peteghem_experimental_2014} metric. Three levels of precedence complexity are examined, with order strength values of 0.75, 0.5, and 0.25, corresponding to high, medium, and low precedence density, respectively. The resulting test scenarios are denoted as 0.75/R12, 0.5/R12, and 0.25/R12. To obtain reliable performance estimates, each GP individual is evaluated on five project instances, and the average makespan relative to the corresponding lower bound is used as the fitness value. 

The proposed algorithm is referred to as SKGGP, i.e., surrogate-assisted KGGP. For the PC scheme, ten decision situations are sampled from the simulation runtime for each type of decision situation. Each sampled situation contains more than ten eligible activity–mode pairs or activity–mode groups. To investigate the impact of different offspring multipliers 
$k$ on the evolutionary process, four values are considered: $k=1, 1.5, 2 \text{ and } 4$. The corresponding algorithms are denoted as SKGGP-1, SKGGP-1.5, SKGGP-2, and SKGGP-4, respectively. All GP algorithms are implemented using the Python DEAP \footnote{https://github.com/DEAP/deap/} framework, with a population size of 1000 and 100 generations, resulting in up to $10^5$ full fitness evaluations. Each algorithm is independently executed 30 times for each scenario. All experiments were conducted on identical computational nodes within a high-performance computing cluster -- New Zealand eScience Infrastructure (NeSI) \footnote{https://www.nesi.org.nz/}.

\subsection{Test Performance}

\begin{figure}
    \centering
    \includegraphics[width=\linewidth]{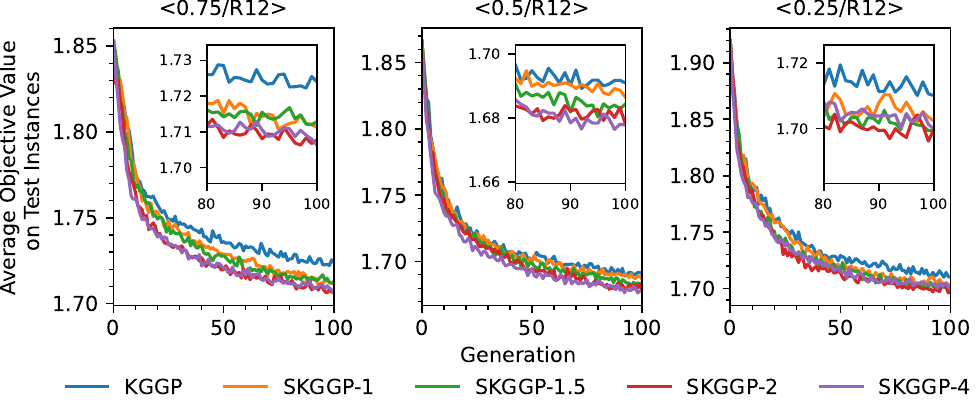}
    \caption{Convergence curves of five algorithms over 30 independent runs.}
    \label{fig:SGP_rank_diff_convergece_curve}
\end{figure}

Table \ref{tab:test_performance} reports the mean and standard deviation of the test performance achieved by the heuristic rules trained using KGGP (baseline) and SKGGP with different offspring multipliers $k$ across the three scenarios. Statistical significance between algorithms is examined using the Wilcoxon signed-rank test. In the table, each algorithm is compared with the ones above it under the same scenario, and the results are indicated using the symbols “($\uparrow$)” (significantly better), “($\downarrow$)” (significantly worse), and “($\approx$)” (no significant difference).
Fig. \ref{fig:SGP_rank_diff_convergece_curve} illustrates the performance of the best individual from each generation, evaluated on the test set during the training process.

Overall, SKGGP demonstrates superior performance to KGGP when a larger number of intermediate offspring are generated. In particular, SKGGP-2 and SKGGP-4 outperform KGGP across all scenarios. The convergence curves further show that SKGGP-2 and SKGGP-4 converge substantially faster than KGGP. Although SKGGP-4 generates twice as many intermediate offspring per generation as SKGGP-2, the performance difference between the two variants is relatively small. The impact of different offspring multipliers on offspring selection is further analysed in Section \ref{section:surrogate_performance}.
For cases with fewer intermediate offspring, such as SKGGP-1.5, improved performance over KGGP is observed in two of the three scenarios. The above comparisons verify the effectiveness of the proposed PC scheme and the surrogate model design.
Notably, SKGGP-1, which does not generate additional intermediate offspring, converges faster than KGGP in the 0.75/R12-1 scenario. This result indicates that the PC-based duplicate removal mechanism alone can also contribute to performance improvement to some extent.


\begin{table}
\centering
\caption{ The mean (standard deviation) of the objective values obtained by five algorithms from 30
independent runs.}
\label{tab:test_performance}
\begin{tabular}{llll}
\toprule
Algorithm & <0.75/R12> & <0.5/R12> & <0.25/R12> \\
\midrule
KGGP & 1.724 ± 0.013 & 1.691 ± 0.011 & 1.710 ± 0.016 \\
SKGGP-1 & 1.711 ± 0.010 & 1.686 ± 0.014 & 1.702 ± 0.014 \\
& ($\uparrow$) & ($\approx$) & ($\approx$) \\
SKGGP-1.5 & 1.713 ± 0.013 & 1.684 ± 0.016 & 1.699 ± 0.014 \\
& ($\uparrow$)($\approx$) & ($\approx$)($\approx$) & ($\uparrow$)($\approx$) \\
SKGGP-2 & 1.708 ± 0.009 & 1.678 ± 0.014& 1.700 ± 0.015 \\
& ($\uparrow$)($\approx$)($\approx$) &($\uparrow$)($\uparrow$)($\approx$)  & ($\uparrow$)($\approx$)($\approx$) \\
SKGGP-4 & 1.706±0.010 & 1.678±0.014 & 1.700±0.021 \\
& ($\uparrow$)($\uparrow$)($\uparrow$)($\approx$) & ($\uparrow$)($\uparrow$)($\uparrow$)($\approx$) & ($\uparrow$)($\approx$)($\approx$)($\approx$) \\
\bottomrule
\end{tabular}
\end{table}

\subsection{Saved Budget Analysis}

\begin{figure}
    \centering
    \includegraphics[width=\linewidth]{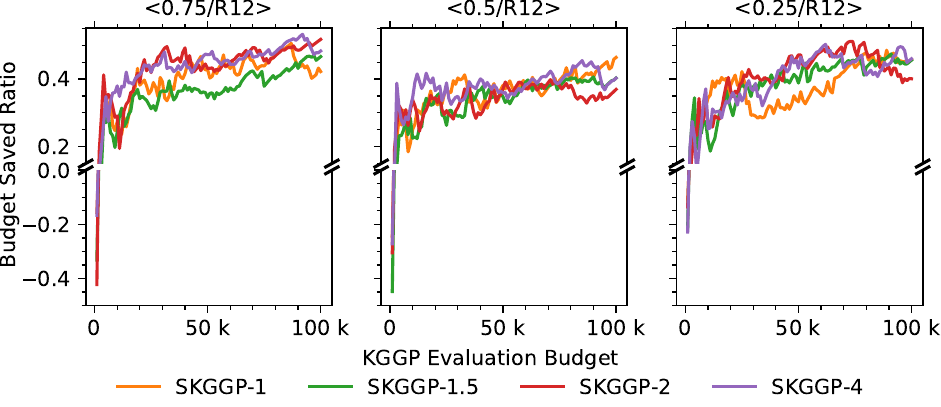}
    \caption{Budget saved to reach the performance of KGGP.}
    \label{fig:SGP_rank_diff_budget_saved}
\end{figure}

Figure \ref{fig:SGP_rank_diff_budget_saved} provides an alternative perspective by illustrating the ratio of full fitness evaluations saved while still achieving the same performance as the baseline KGGP. The x-axis represents the number of full fitness evaluations already consumed by the baseline KGGP. For a given x value, the y-axis shows the budget saved ratio, which is computed by first identifying the best solution found by KGGP at that evaluation count, and then locating the earliest evaluation at which another algorithm discovers a solution of equal or better quality. The difference between these two evaluation indices, normalised by the number of evaluations used by KGGP, defines the budget saved ratio. 

This analysis highlights the ability of surrogate-assisted KGGP (SKGGP) to identify heuristic rules of comparable quality using fewer full evaluations and at earlier stages of evolution.
In the early phase of evolution, the ratio is negative, indicating that the heuristic rules discovered by SKGGP are inferior to those found by KGGP at the same number of evaluations. However, as evolution progresses, SKGGP begins to accelerate its convergence. Once heuristic rules of equivalent quality to those of KGGP are reached, SKGGP can save approximately 20–40\% of full fitness evaluations.
These results demonstrate that the surrogate model effectively pre-selects intermediate offspring, thereby improving the quality of individuals entering the next generation. As a consequence, high-quality heuristic rules can be discovered earlier, leading to a more evaluation-efficient evolutionary process.

\subsection{Surrogate Performance and Impact of Offspring Multiplier on Offspring Quality}
\label{section:surrogate_performance}

To further investigate the performance of the surrogate model and the impact of different offspring multipliers $k$ on the evolutionary process, we conduct an additional analysis based on the population generated by the baseline KGGP at each generation. Specifically, for each generation, four times the population size of intermediate offspring are generated, and their true fitness values are obtained through full evaluation. These true fitness values are then compared with the fitness values estimated by the surrogate model.

This analysis aims to address the following two questions:
\begin{itemize}
    \item Whether the surrogate model is capable of distinguishing high-quality individuals from low-quality ones.
    \item To what extent increasing the number of intermediate offspring improves offspring quality.
\end{itemize}

\begin{figure}
    \centering
    \includegraphics[width=\linewidth]{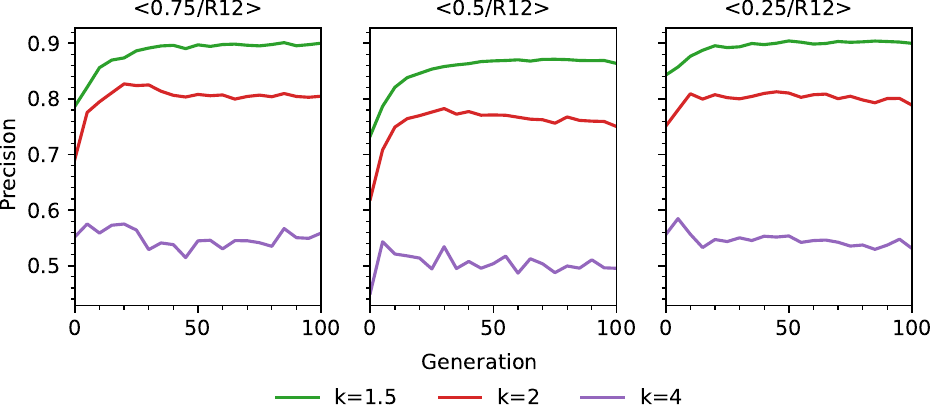}
    \caption{Mean precision by various offspring multipliers $k$ across generation.}
    \label{fig:surrogate_precision}
\end{figure}

To answer the first question, we measure the precision of the surrogate model at each generation under different values of $k$, and the results are shown in Fig. \ref{fig:surrogate_precision}. This metric reflects the extent to which the surrogate model can correctly select individuals that should be retained for the next generation.
Here, a true positive is defined as an individual whose true fitness ranks within the top 1000 and whose estimated fitness also ranks within the top 1000. A false positive refers to an individual whose estimated fitness ranks within the top 1000, but whose true fitness rank is worse than 1000. The precision is the ratio of true positive instances over all positive instances. 

The precision of the surrogate model decreases as the offspring multiplier $k$ increases. When $k=1.5$, the precision reaches approximately 80–90\%. For k=2, the precision drops to below 70\%, while for $k=4$, it further decreases to around 50–60\%. This trend is intuitive, as selecting the top 1000 individuals becomes increasingly challenging as the number of intermediate offspring grows.
One possible explanation for this phenomenon lies in the limited size of the surrogate database. Since the database only contains PC vectors from the current population, the surrogate model may lack sufficient representative samples to make accurate predictions when faced with a large number of candidate offspring.

\begin{figure}
    \centering
    \includegraphics[width=\linewidth]{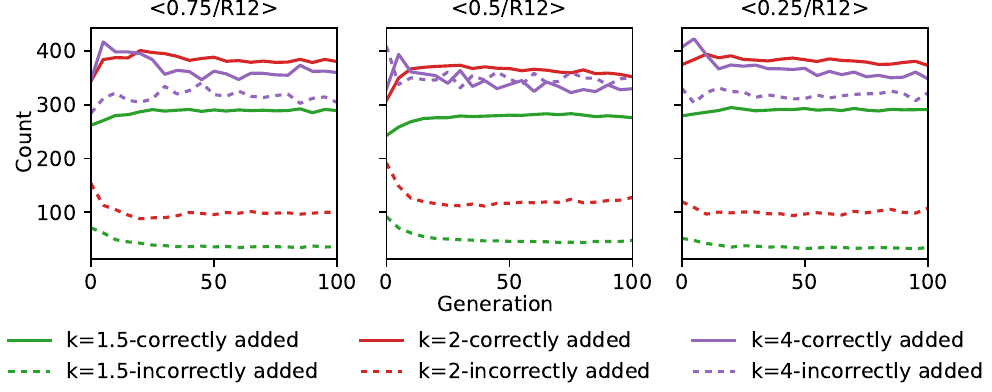}
    \caption{Mean count of true positive and false positive by various offspring multipliers $k$ across generations.}
    \label{fig:surrogate_extra_gain}
\end{figure}

To answer the second question, the intermediate offspring are further divided into two groups: base offspring, which consist of the first 1000 generated offspring, and extra offspring, which include the offspring generated beyond the population size (i.e., the 1001st and subsequent offspring). We then examine how many individuals from the extra offspring are selected into the top 1000 based on estimated fitness.
Among these selected extra offspring, individuals whose true fitness and estimated fitness both rank within the top 1000 are classified as correctly added, while those whose estimated fitness ranks within the top 1000 but whose true fitness ranks outside the top 1000 are classified as incorrectly added. The corresponding statistics are reported in Fig. \ref{fig:surrogate_extra_gain}. 

When $k=1.5$, approximately 300 individuals per generation are selected from the extra offspring (1001–1500), most of which are correctly added, with only about 50 individuals being incorrectly added. For $k=2$, around 350 individuals are correctly added, while approximately 100 individuals are incorrectly added. This indicates that, for $k=2$, about 35\% of the extra offspring successfully replace the original offspring, enter the subsequent evolutionary process and contribute to improving population quality. This observation further demonstrates the effectiveness of generating intermediate offspring followed by surrogate-based selection.

When $k=4$, although the number of correctly added individuals slightly increases compared to $k=2$, the number of incorrectly added individuals rises substantially to around 300. This result is consistent with the lower precision observed for $k=4$ and partially explains why SKGGP-2 and SKGGP-4 achieve similar overall performance. Although SKGGP-4 generates twice as many intermediate offspring per generation as SKGGP-2, the surrogate model is less capable of accurately distinguishing high-quality individuals from low-quality ones, leading to the introduction of both beneficial and detrimental individuals into the population.

\subsection{Surrogate Overhead}

To examine whether the introduction of the surrogate model imposes a significant computational overhead, we report the average per-generation evaluation time of SKGGP across different scenarios in Table \ref{tab:evaluation_comparison}. Full evaluation refers to the time required to evaluate $|P|$ individuals (i.e., 1000) in the current generation using the simulation model. Surrogate estimation denotes the time needed to compute the PC vectors of $k*|P|$ intermediate offspring and to estimate their fitness values using the surrogate model. Overall, the time spent on surrogate estimation is approximately 1/20 to 1/40 of that required for full evaluation. This indicates that the surrogate-related overhead is relatively small, demonstrating that the surrogate model can enhance the convergence of the algorithm with only a marginal increase in computational cost.


\begin{table}[]
    \centering
    \caption{Mean evaluation time of full evaluation and surrogate estimation in seconds.}
    \label{tab:evaluation_comparison}
    \begin{tabular}{llrr}
    \toprule
     Scenario &  Algorithm & Full Evaluation & Surrogate Estimation \\
     \midrule
    \multirow[t]{4}{*}{0.75/R12} & SKGGP-1 & 266.27 & 9.65 \\
     & SKGGP-1.5 & 275.01 & 10.94 \\
     & SKGGP-2 & 333.73 & 13.05 \\
     & SKGGP-4 & 362.33 & 20.26 \\
    \multirow[t]{4}{*}{0.5/R12} & SKGGP-1 & 556.13 & 14.00 \\
     & SKGGP-1.5 & 602.91 & 19.63 \\
     & SKGGP-2 & 541.95 & 22.89 \\
     & SKGGP-4 & 593.04 & 45.40 \\
    \multirow[t]{4}{*}{0.25/R12} & SKGGP-1 & 784.37 & 10.53 \\
     & SKGGP-1.5 & 967.90 & 13.83 \\
     & SKGGP-2 & 948.97 & 15.27 \\
     & SKGGP-4 & 881.96 & 23.98 \\
     \bottomrule
    \end{tabular}
\end{table}
\section{Conclusions}

This paper proposes a phenotypic characterisation scheme for genetic programming applied to the dynamic multi-mode resource-constrained project scheduling problem. By analysing the decision situations encountered during GP-based scheduling, a rank-based PC vector is designed to transform the behavioural characteristics of GP individuals into a numerical vector. This addresses the lack of suitable PC schemes for GP in DMRCPSP and enables behaviour-level comparison between individuals.
The proposed PC scheme is integrated into a surrogate-assisted GP framework, where it is used to measure distances between GP individuals and to estimate the fitness of unevaluated intermediate offspring. Experimental results demonstrate that the SKGGP algorithm is able to evolve higher-quality heuristic rules than the baseline KGGP with significantly fewer expensive fitness evaluations. Further analyses reveal the surrogate model’s ability to pre-select promising offspring and quantify the contribution of different numbers of intermediate offspring to population quality.

Despite these advantages, the precision of the current surrogate model decreases when selecting from a large number of intermediate offspring. Future work will therefore focus on improving surrogate accuracy by expanding the surrogate database, exploring more expressive PC schemes, and investigating alternative machine learning models for fitness estimation.

\bibliography{references}

\end{document}